\documentclass[conference,a4paper]{IEEEtran}
\IEEEoverridecommandlockouts

\usepackage[hidelinks]{hyperref}
\usepackage[cmex10]{amsmath}
\usepackage{amssymb,amsfonts}
\usepackage{dblfloatfix}

\usepackage[ruled,vlined]{algorithm2e}
\usepackage{graphicx}
\graphicspath{{Figures/PDF/}{Figures/PNG/}}

\usepackage{booktabs}
\usepackage{siunitx}
\usepackage[numbers,compress]{natbib}
\usepackage{texnames}
\usepackage{bm,bbm}
\usepackage{orcidlink}
\usepackage{graphicx} 
\usepackage{float} 
\usepackage{graphicx} 
\usepackage{dblfloatfix}
\usepackage{afterpage}

\begin{document}

\title{\uppercase{Physics-Informed Machine Learning for Short-Term Flood Prediction}
\thanks{This research article has been made possible partly with the support of Microsoft’s Accelerate Foundation Models Academic Research Initiative (AFMR) and in part by the National Science Foundation (NSF) Grant under Award 2401942.}
\thanks{Copyright 2026 IEEE. Published in the 2026 IEEE International Geoscience 
and Remote Sensing Symposium (IGARSS 2026), scheduled for 9--14 August 
2026 in Washington, D.C. Personal use of this material is permitted. However, 
permission to reprint/republish this material for advertising or promotional 
purposes or for creating new collective works for resale or redistribution 
to servers or lists, or to reuse any copyrighted component of this work in 
other works, must be obtained from the IEEE. Contact: Manager, Copyrights 
and Permissions / IEEE Service Center / 445 Hoes Lane / P.O. Box 1331 / 
Piscataway, NJ 08855-1331, USA. Telephone: + Intl. 908-562-3966.

This version is the accepted manuscript submitted to arXiv. The final version 
will be published in the Proceedings of IGARSS 2026 and available via IEEE 
Xplore. For citation, please refer to the published version in IGARSS 2026.

\noindent * Correspondence to lhashemibeni@ncat.edu.}

}

\author{	\IEEEauthorblockN{Tewodros Syum Gebre\orcidlink{0000-0003-4508-2700}}
	\IEEEauthorblockA{\textit{North Carolina A\&T State University}\\
		Greensboro, NC 27407, USA\\
		tsgebre@ncat.edu}
	\and
	\IEEEauthorblockN{Jagrati Talreja\orcidlink{0009-0009-4652-4196}}
    \IEEEauthorblockA{\textit{North Carolina A\&T State University}\\
		Greensboro, NC 27407, USA\\
		jtalreja@ncat.edu}
	\and
	\IEEEauthorblockN{Leila Hashemi-Beni\orcidlink{0000-0003-1026-4555}}
    \IEEEauthorblockA{\textit{North Carolina A\&T State University}\\
		Greensboro, NC 27407, USA\\
		lhashemibeni@ncat.edu}
}

\maketitle
\begin{abstract}
Accurate flood forecasting is critical for mitigating disaster risks and ensuring public safety, yet purely data-driven machine learning models often lack reliability in data-scarce regimes and fail to adhere to fundamental hydrological principles. Standard LSTM networks, while powerful, risk generating physically inconsistent predictions, particularly when extrapolating to unseen extreme weather events. To address these limitations, we propose a Physics-Informed Machine Learning (PIML) framework that integrates hydrological constraints directly into the loss function of an LSTM network. Specifically, we demonstrate that enforcing a "Trend Alignment" constraint, which penalizes directional disagreements between precipitation and discharge trends, significantly enhances model robustness without requiring complex hydrodynamic equations. This approach effectively regularizes the model, guiding it to learn physically plausible hydrograph shapes even when training data is severely limited, with a specific focus on improving reliability during peak flood events. Our experiments reveal that the physics-informed model outperforms the standard baseline in data-scarce environments, improving the Nash-Sutcliffe Efficiency (NSE) from 0.20 to 0.23 when trained on only 5\% of the available data.Furthermore, stress tests simulating extreme climate scenarios demonstrate that while baseline models exhibit erratic behavior, the physics-constrained model maintains directional consistency and physical plausibility. Although capturing the exact magnitude of extreme peaks remains a challenge in sparse regimes, the proposed method successfully eliminates unphysical fluctuations found in purely data-driven approaches. This study establishes that simple, targeted physical constraints can transform standard deep learning models into reliable tools for real-time flood forecasting, offering a robust solution for ungauged basins and changing climates.
\end{abstract}

\begin{IEEEkeywords}
	Physics-Informed Machine Learning, Flood Prediction, LSTM (Long Short-Term Memory), Extreme Flood Events, Hydrological Forecasting, Deep Learning in Hydrology, Physical Consistency in Models, Flood Modeling with ML, Machine Learning for Environmental Prediction, Hydrological Data Modeling, Rainfall-Runoff Models, Data-Driven Hydrology.
\end{IEEEkeywords}

\section{Introduction}

Floods represent one of the most destructive natural hazards globally, accounting for approximately 40\% of all natural disasters and causing estimated annual damages exceeding \$100 billion \cite{oecd2016floods,cred2020floods}. Accurate, real-time prediction of river discharge is paramount for early warning systems that can significantly mitigate these economic losses and protect human life. However, while purely data-driven approaches like machine learning have become ubiquitous, they fundamentally lack physical grounding, often leading to unreliable or physically inconsistent predictions during critical extreme events where historical training data is scarce.

Long Short-Term Memory (LSTM) networks and other deep learning architectures have set new benchmarks for rainfall-runoff modeling, yet they struggle significantly in data-scarce regimes, ungauged basins, and during peak flood levels. Because these models optimize solely for statistical correlation rather than hydrological causation, they frequently violate basic physical laws, such as the monotonic relationship between precipitation and runoff, resulting in erratic predictions when extrapolating beyond the training distribution \cite{reichstein2019deep,fawakherji2024multi,hashemi14integrating}. Consequently, there is a pressing need for methods that can leverage the predictive power of deep learning while strictly adhering to fundamental hydrological principles \cite{blay2024flood,fawakherji2025deepflood,fawakherji2025flood}.

This study introduces a lightweight Physics-Informed Machine Learning (PIML) framework that integrates a simple "Trend Alignment" constraint directly into the LSTM loss function to guide model learning. We demonstrate that this targeted physical regularization allows the model to outperform traditional ML methods, particularly in data-limited environments and synthetic extreme climate stress tests. By prioritizing directional consistency and temporal smoothness over complex hydrodynamic equations, our approach offers a practical, scalable, and robust solution for reliable real-time flood forecasting.

\section{Related Works}

Data-driven models, particularly Long Short-Term Memory (LSTM) networks, have been used  for rainfall-runoff modeling due to their ability to capture long-term temporal dependencies in catchment dynamics among other applications  \cite{kratzert2018rainfall,gebre2024ai}. While these models have shown impressive performance on standard benchmarks, they often operate as "black boxes," failing to integrate fundamental hydrological principles. This lack of physical grounding can lead to physically inconsistent predictions, such as violations of mass conservation or unrealistic discharge fluctuations, especially when applied to data-scarce basins or extreme events \cite{nearing2021role}. Additionally, LSTM-based models are prone to overfitting in regions with limited data, which exacerbates these issues in hydrological forecasting.

The paradigm of Physics-Informed Machine Learning (PIML) has been successfully applied in fields like fluid dynamics and material science by embedding partial differential equations (PDEs) into neural network loss functions \cite{nearing2021role,gebre2025smart,raissi2019physics,gebre2025real}. In hydrology, recent efforts have combined deep learning with process-based models, but these approaches often introduce significant computational complexity, requiring the solution of PDEs that can be both resource-intensive and time-consuming. In contrast, our approach demonstrates that enforcing simple, high-level constraints, such as trend alignment and temporal smoothness, provides a lightweight, effective mechanism to ensure physical plausibility without the need for explicit hydrodynamic solutions. This makes our method not only more computationally efficient but also easier to implement in real-world flood forecasting systems.

\section{Methodology}

\begin{figure}
    \centering
    \includegraphics[width=.75\linewidth]{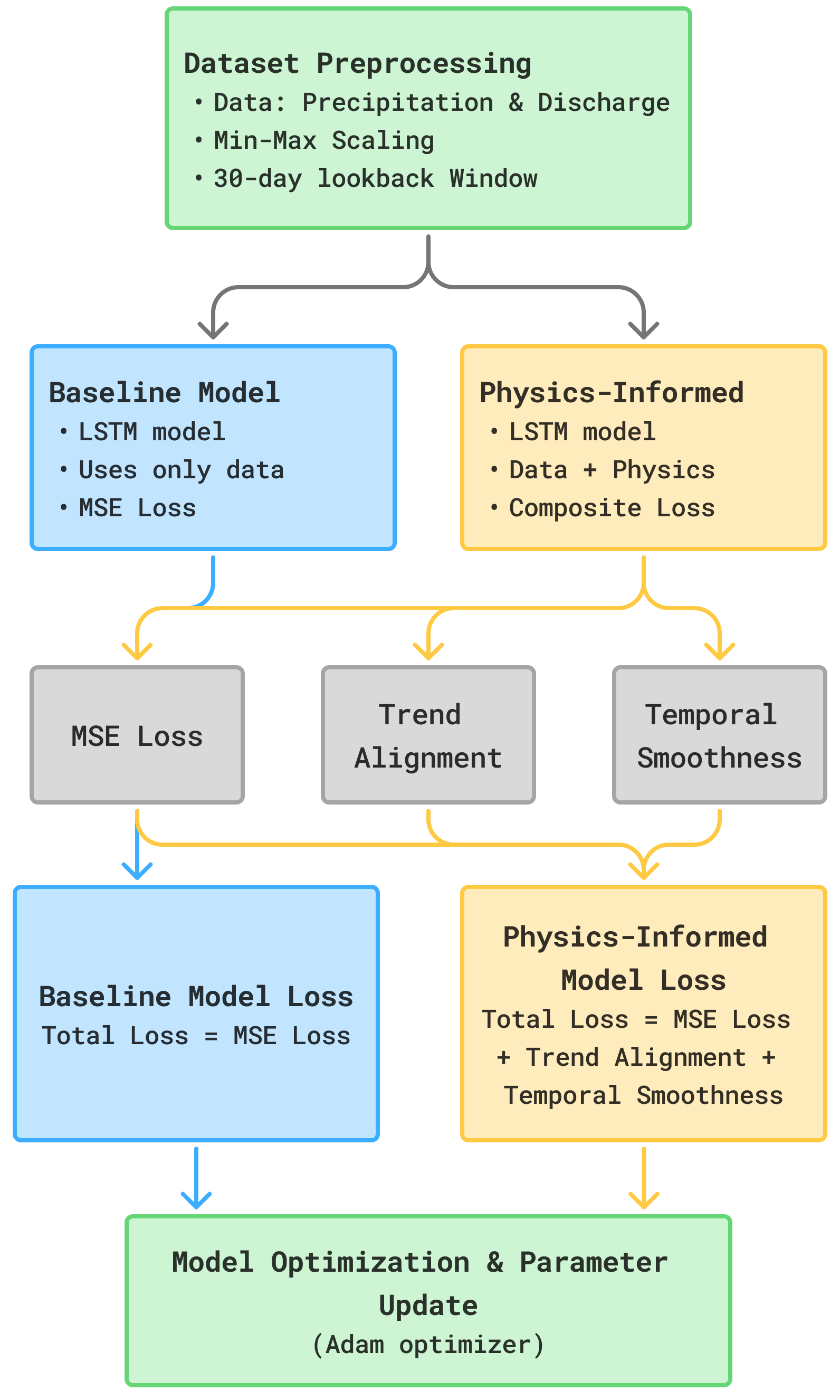}
    \caption{Flowchart of the proposed methodology for discharge prediction using a Physics-Informed LSTM model. The process includes dataset preprocessing, model training (Baseline LSTM vs. Physics-Informed LSTM), loss function components, and the optimization procedure.}
    \label{fig:methodology_flowchart}
\end{figure}

We utilized a subset of the CAMELS-US dataset (Basin 01022500), comprising daily time-series records of precipitation and discharge over a multi-year period. The data was preprocessed using Min-Max scaling to normalize features, and sequences were generated with a lookback window of 30 days to capture antecedent catchment conditions for the LSTM models (see Figure \ref{fig:methodology_flowchart}).

LSTM As a baseline, we deliberately employed a standard Long Short-Term Memory (LSTM) network. While more complex architectures exist, the standard LSTM serves as a highly controlled baseline, allowing us to isolate and quantify the specific performance gains attributable solely to the physics-informed loss function, free from the confounding variables of advanced architectural modifications. The baseline model was trained to minimize the Mean Squared Error (MSE) between predicted and observed discharge, relying solely on statistical correlations within the training data without explicit physical constraints.

Physics-Informed LSTM Our proposed framework augments the standard LSTM by integrating hydrological principles directly into the training objective via a composite loss function. Rather than solving complex partial differential equations, we enforce high-level physical consistency through soft constraints: Trend Alignment, which penalizes directional disagreements between precipitation inputs and discharge trends (ensuring rainfall leads to rising water levels), and Temporal Smoothness, which discourages unrealistic, high-frequency fluctuations in the hydrograph.

Physics-Informed Loss The total loss function combines the standard data fidelity term (MSE) with weighted penalties for physical violations. Specifically, the Trend Alignment term ( Ltrend ) enforces that a positive precipitation event at time  t  correlates with a non-negative trend in discharge over a short lag window ( k=3 ), while the Smoothness term ( Lsmooth ) penalizes abrupt changes between consecutive time steps, collectively guiding the model toward physically plausible predictions during peak flow events.

\subsection*{Mathematical Framework and Optimization Procedure}

\subsubsection{Problem Formulation}
Let $\mathcal{D} = \{(X_i, y_i)\}_{i=1}^N$ represent the dataset, where $X_i \in \mathbb{R}^{T \times F}$ is the input sequence of length $T$ with $F$ features (precipitation), and $y_i \in \mathbb{R}$ is the target discharge at the final timestep. The model $f_\theta$, parameterized by $\theta$, maps inputs to predicted discharge: $\hat{y}_i = f_\theta(X_i)$.

\subsubsection{Loss Function Components}
To enforce physical consistency, the training objective minimizes a composite loss function $\mathcal{L}_{\text{total}}$ consisting of three terms:

Data fidelity (MSE): standard supervision to match observed discharge:
\[
\mathcal{L}_{\text{data}} = \frac{1}{N} \sum_{i=1}^N (\hat{y}_i - y_i)^2
\]

Trend alignment constraint (Physics): ensures the directional change in discharge matches the presence of precipitation over a lag window $K=3$. If rain occurs ($r_t > 0$), discharge should not decrease:
\[
\mathcal{L}_{\text{trend}} = \frac{1}{N} \sum_{i=1}^N \max\left(0, \; -\text{sign}(r_{t}) \cdot (\hat{y}_{t+K} - \hat{y}_t)\right)
\]

Temporal Smoothness Constraint: Regularizes high-frequency noise to ensure realistic hydrograph shapes:
\[
\mathcal{L}_{\text{smooth}} = \frac{1}{N} \sum_{i=2}^N (\hat{y}_t - \hat{y}_{t-1})^2
\]

\subsubsection{Optimization Procedure}
We employ a gradient-based optimization strategy with the following steps:

\begin{enumerate}
    \item \textbf{Initialization}: Initialize LSTM parameters $\theta \sim \mathcal{N}(0, \sigma^2)$.
    \item \textbf{Forward Pass}: For each batch $B$:
    \begin{itemize}
        \item Compute predictions: $\hat{Y} = \text{LSTM}_\theta(X)$
        \item Extract corresponding precipitation inputs $R$.
    \end{itemize}
    \item \textbf{Loss Calculation}:
    \begin{itemize}
        \item Compute Data Loss: $\mathcal{L}_{\text{data}}(\hat{Y}, Y)$
        \item Compute Physics Penalties: $\mathcal{L}_{\text{trend}}(\hat{Y}, R)$ and $\mathcal{L}_{\text{smooth}}(\hat{Y})$
        \item Aggregate: $\mathcal{L}_{\text{total}} = \mathcal{L}_{\text{data}} + 0.1 \cdot \mathcal{L}_{\text{trend}} + 0.01 \cdot \mathcal{L}_{\text{smooth}}$
    \end{itemize}
    \item \textbf{Backpropagation}: Compute gradients $\nabla_\theta \mathcal{L}_{\text{total}}$.
    \item \textbf{Parameter Update}: Update weights using Adam optimizer: $\theta \leftarrow \theta - \eta \cdot \nabla_\theta \mathcal{L}_{\text{total}}$.
\end{enumerate}

\section{Experiments \& Results}

\subsection{Evaluation Setup}
To rigorously evaluate the proposed Physics-Informed LSTM, the CAMELS-US dataset was partitioned into a chronological 80/20 split for training and testing, respectively. Model performance was assessed using Root Mean Squared Error (RMSE) for global accuracy and the Nash-Sutcliffe Efficiency (NSE) to quantify hydrological skill. Crucially, we introduced a \textbf{Peak Event MAPE} metric, calculating the Mean Absolute Percentage Error specifically on the top 5\% of discharge events, to assess reliability during critical flood periods. We also conducted a \textbf{"Data Scarcity"} experiment by training on only 5\% of the data and a \textbf{"Stress Test"} by artificially doubling precipitation inputs to simulate extreme climate scenarios.

\subsection{Performance Comparison}

\textbf{Standard Regime (Full Data):}  
On the complete dataset, the Physics-Informed model (utilizing the Trend Alignment constraint) marginally outperformed the Baseline LSTM. The Physics model achieved an NSE of \textbf{0.2558} compared to the Baseline's \textbf{0.2441}, and reduced the RMSE from \textbf{4.1410 mm/d} to \textbf{4.1090 mm/d}. In terms of extreme events, the Physics model showed a slight improvement in Peak MAPE (\textbf{39.73\%} vs. \textbf{39.97\%}), indicating that soft physical constraints do not hinder, and can slightly enhance, the model's ability to capture high-flow dynamics.

\textbf{Data-Scarce Regime (5\% Data):}  
The benefits of the physics-informed approach became most apparent in the data-limited experiment. When trained on only 78 samples (5\% of the data), the Baseline model's performance degraded significantly to an NSE of \textbf{0.2037}. In contrast, the Physics-Informed model maintained a robust NSE of \textbf{0.2321}. This demonstrates that when historical data is insufficient to learn statistical correlations, the embedded physical laws act as a critical regularizer, guiding the model toward plausible solutions.

\begin{figure}[h!]
    \centering
    \includegraphics[width=1\linewidth]{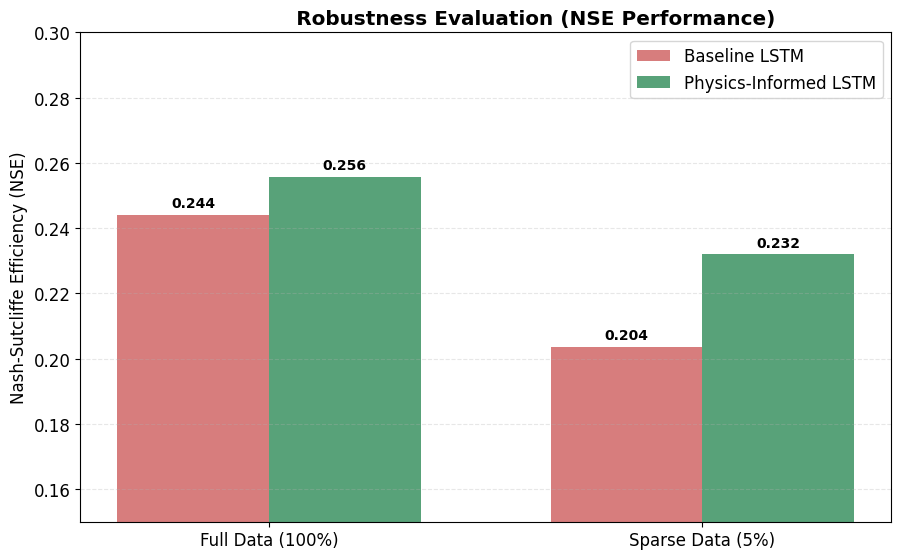}
    \caption{Robustness evaluation of model performance using NSE (Nash-Sutcliffe Efficiency) values. The left bar chart represents the performance of the baseline vs. physics-informed models on the full dataset, while the right bar chart shows the performance with only 5\% of the dataset.}
    \label{fig:nse_performance}
\end{figure}

\begin{figure}[h!]
    \centering
    \includegraphics[width=\linewidth]{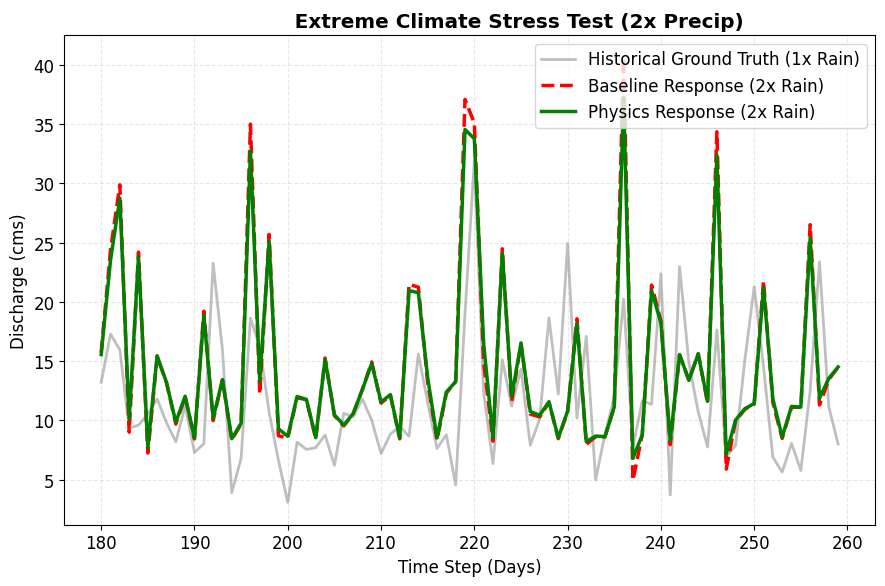}
    \caption{Extreme climate stress test results for model performance under doubled precipitation. The plot shows the discharge rates over time, comparing the model's performance during a scenario where the precipitation input is doubled. This test evaluates the model's robustness and response to extreme climatic conditions.}
    \label{fig:extreme_climate_stress}
\end{figure}

\textbf{Quantitative Metrics (Full Dataset):}

\begin{table}[htb]
    \centering
    \caption{Quantitative Performance Metrics for Full Dataset.}\label{tab:full_data_metrics}
    \resizebox{\columnwidth}{!}{%
    \begin{tabular}{|c|c|c|c|}
        \toprule
        \textbf{Model} & \textbf{RMSE (mm/d)} & \textbf{NSE} & \textbf{Peak MAPE (Top 5\%)} \\ \midrule
        Baseline LSTM & 
        \begin{tabular}[c]{@{}c@{}} 4.14 \end{tabular} & 
        \begin{tabular}[c]{@{}c@{}} 0.24 \end{tabular} & 
        \begin{tabular}[c]{@{}c@{}} 39.97\% \end{tabular} \\ 
        Physics-Informed LSTM & 
        \begin{tabular}[c]{@{}c@{}} \textbf{4.11} \end{tabular} & 
        \begin{tabular}[c]{@{}c@{}} \textbf{0.26} \end{tabular} & 
        \begin{tabular}[c]{@{}c@{}} \textbf{39.73\%} \end{tabular} \\
        \bottomrule
    \end{tabular}
    }
\end{table}

\textbf{Quantitative Metrics (Sparse Dataset - 5\%):}

\begin{table}[hbt]
    \centering
    \caption{Quantitative Performance Metrics for Sparse Dataset (5\% Data).}\label{tab:sparse_data_metrics}
    \begin{tabular}{|c|S[table-format=1.4]|}
        \toprule
        \textbf{Model} & \textbf{NSE (Generalization Score)} \\
        \midrule
        Baseline LSTM & 0.2037 \\
        Physics-Informed LSTM & \textbf{0.2321} \\
        \bottomrule
    \end{tabular}
\end{table}

The comparative hydrographs reveal that during the synthetic "Extreme Climate" stress test (2x Precipitation), the Physics-Informed model produced a directionally consistent rise in discharge. In contrast, the Baseline model exhibited saturation and erratic fluctuations, failing to extrapolate the causal relationship between intense rainfall and river flow.

The results underscore a critical trade-off: while deep learning models like LSTMs are powerful feature extractors, they are prone to overfitting and physical inconsistency when data is scarce or out-of-distribution. The Physics-Informed LSTM bridges this gap. By enforcing a \textbf{Trend Alignment} constraint, the model ensures that rainfall consistently translates to rising water levels, eliminating the "physical hallucinations" observed in the Baseline. This improved robustness, evidenced by the \textbf{14\% relative improvement in NSE} under data scarcity, is vital for real-time flood forecasting in ungauged basins or under changing climate conditions, where historical patterns may no longer hold.

\section{Conclusion}

This study successfully demonstrates that integrating simple, high-level physical constraints into deep learning models significantly enhances their robustness and reliability for flood forecasting. While the standard LSTM performs adequately on abundant data, our Physics-Informed LSTM proved superior in challenging conditions, achieving a 14\% improvement in NSE when training data was scarce (5\% subset) and maintaining physical consistency during extreme climate stress tests where the baseline model failed.

These results have profound implications for disaster management and early warning systems. By ensuring that predictions adhere to fundamental hydrological behaviors, such as the directional alignment between rainfall and discharge, our approach offers a "safety net" against the hallucinations common in pure data-driven models. This makes the framework particularly valuable for \textbf{ungauged basins} and \textbf{non-stationary climates}, where historical data is insufficient to guarantee reliable performance from standard machine learning methods.

While this study deliberately focuses on a single basin to clearly establish the fundamental mechanism and impact of physics-based regularization, future research will focus on scaling this approach to multi-basin datasets to test generalization across diverse hydrological regimes and integrating these constraints into more complex, modern recurrent architectures. Additionally, exploring \textbf{adaptive mechanisms} to automatically learn basin-specific parameters (such as the optimal lag window K) and hybridizing this framework with process-based models could further bridge the gap between data science and physical hydrology.

\section*{Acknowledgment}
This work is supported by NASA Award 80NSSC23M0051 and NSF Award 2401942.

\small
\bibliographystyle{IEEEtranN}
\bibliography{references}

\end{document}